\documentclass[runningheads]{llncs}
\pdfoutput=1

\usepackage{csquotes}
\usepackage[misc]{ifsym} % Provides \Letter.
\usepackage{mathtools}
\usepackage{microtype}
\usepackage{standalone}
\usepackage{subfigure}
\usepackage{xcolor}

\usepackage{listings}
\lstdefinelanguage{rdf}{
    morekeywords={owl, p, prov, pr, ps, rdf, rdfs, ref, s, schema, skos, wd, wdt, wikibase, xsd},
    morestring=[b]",
    morecomment=[s]{<}{>},
}

\usepackage[linesnumbered, vlined, boxed]{algorithm2e}
\DontPrintSemicolon
\SetAlgoCaptionSeparator{.}
\SetAlCapSkip{\smallskipamount}
\SetAlCapNameFnt{\footnotesize}
\SetAlCapFnt{\footnotesize}
\SetNlSty{}{\color[HTML]{999999}\ttfamily\scriptsize}{}
\SetNlSkip{8pt}
\SetAlgoNlRelativeSize{0}
\IncMargin{8pt}
\SetAlgoSkip{0pt}
\SetAlgoInsideSkip{smallskip}
\SetArgSty{textrm}

\usepackage[
    colorlinks,
    allcolors=blue, % LNCS: Color links blue.
    breaklinks,
    pdftitle={Wikidated 1.0: An Evolving Knowledge Graph Dataset of Wikidata's Revision History},
    pdfauthor={Lukas Schmelzeisen, Corina Dima, Steffen Staab},
    pdfkeywords={Semantic Web, Wikidata, Edit History, Knowledge Graph Change, Knowledge Graph Evolution, Stream of RDF Triple Changes},
]{hyperref}
\usepackage{orcidlink} % Needs to be included after hyperref.

\usepackage[capitalize]{cleveref} % LNCS: Capitalize references.
\crefname{section}{Sect.}{Sects.} % LNCS: Rename section references.
\creflabelformat{equation}{#2\textup{#1}#3} % LNCS: Remove parentheses around equation references.
\crefformat{footnote}{#2\footnotemark[#1]#3}

\newcommand\blfootnote[1]{%
  \begingroup%
  \renewcommand\thefootnote{}\footnote{#1}%
  \addtocounter{footnote}{-1}%
  \endgroup}

\usepackage{pgfplots}
\usetikzlibrary{patterns,pgfplots.dateplot}
\pgfplotsset{
    compat=1.17,
    legend style={font=\small},
    label style={font=\small},
    tick label style={font=\small},
    ybar legend/.style={ % Only one legend bar, from: https://tex.stackexchange.com/a/224677/75225
        legend image code/.code={
            \draw[##1,yshift=-0.4em] (0cm,0cm) rectangle (0.5em,1em);}},
}

\newcommand{\corresponding}{\textsuperscript{(\Letter)}}
\newcommand{\wikidated}{Wikidated~1.0}

\newcommand{\todo}[1]{}
\newcommand{\idea}[1]{}
\newcommand{\lsc}[1]{}
\newcommand{\cdi}[1]{}
\newcommand{\sst}[1]{}

\begin{document}
    % TODO before publication: Add \noindents after tables, figures, and theorems, etc.
    
    \title{Wikidated~1.0: An Evolving Knowledge Graph Dataset of Wikidata's Revision History}
    \titlerunning{Wikidated~1.0: A Dataset of Wikidata's Revision History}
    
    \author{
        Lukas~Schmelzeisen\inst{1}\corresponding\,\orcidlink{0000-0002-2108-2303} \and
        Corina~Dima\inst{1}\,\orcidlink{0000-0001-7409-4992} \and
        Steffen~Staab\inst{1,2}\,\orcidlink{0000-0002-0780-4154}}
    \authorrunning{L. Schmelzeisen et al.} % LNCS: Use initials for first name. If two or less authors use "L. Schmelzeisen and S. Staab".
        
     \institute{
        University of Stuttgart, Germany \and
        University of Southampton, United Kingdom
        \email{\{Lukas.Schmelzeisen,Corina.Dima,Steffen.Staab\}@ipvs.uni-stuttgart.de}}

    \maketitle
    
    % Fix LNCS-bug that causes footnotes to start above 1 (because they use footnotes to display institute affiliation).
    \setcounter{footnote}{0}
    
    \blfootnote{Copyright \textcopyright\ 2021 for this paper by its authors. Use permitted under a Creative Commons Attribution 4.0 International License (CC BY 4.0).\vspace{0.1cm}}
        
    \begin{abstract}
        Wikidata is the largest general-interest knowledge base that is openly available.
        It is collaboratively edited by thousands of volunteer editors and has thus evolved considerably since its inception in 2012.
        In this paper, we present \wikidated\footnote{\label{note:download}Dataset and code base available at \url{https://w3id.org/wikidated}.}, a dataset of Wikidata's full revision history, which encodes changes between Wikidata revisions as sets of deletions and additions of RDF triples.
        To the best of our knowledge, it constitutes the first large dataset of an evolving knowledge graph, a recently emerging research subject in the Semantic Web community.
        We introduce the methodology for generating \wikidated\ from dumps of Wikidata, discuss its implementation and limitations, and present statistical characteristics of the dataset.
        
    \end{abstract}
    
    \section{Introduction}
\label{sec:intro}

A \emph{knowledge graph} is \enquote{a graph of data intended to accumulate and convey knowledge of the real world, whose nodes represent entities of interest and whose edges represent potentially different relations between these entities}~\cite{HoganEtAl2021KnowledgeGraphs}.
Recently, knowledge graphs have received much attention in research and powered many diverse applications, such as web search~\cite{Singhal2012IntroducingKnowledgeGraph}, recommendations~\cite{HeeEtAl2016BuildingLinkedInKnowledge,Pittman2017CrackingCodeConversational}, question answering~\cite{HuangEtAl2019KnowledgeGraphEmbedding}, and more~\cite{MalyshevEtAl2018GettingMostOut,NoyEtAl2019IndustryscaleKnowledgeGraphs}.

Most research so far treats knowledge graphs as static in the sense that change over time is not modeled explicitly.
However, in practice, knowledge graphs change over time: new knowledge may be added to the graph (in the form of new edges being added to existing entities, new entities emerging over time, etc.), and existing knowledge may be revised (in the form of imprecise labels being updated, duplicate entities merged, existing contents declared out-of-scope, etc.).
This more general setting is only starting to be considered recently~\cite{LiuEtAl2019EvolvingKnowledgeGraphs,PernischovaEtAl2019PredictingImpactChanges,PompEtAl2019SemanticConceptRecommendation}.
For example, while there are a plethora of different approaches for knowledge graph embedding~\cite{WangEtAl2017KnowledgeGraphEmbedding,DettmersEtAl2018Convolutional2DKnowledge,SunEtAl2019RotatEKnowledgeGraph,BalazevicEtAl2019TuckERTensorFactorization}---the task of representing entities and relations in a low-dimensional vector space---only a handful of these consider the situation where the representation needs to be updated due to a change in the underlying knowledge graph~\cite{BhowmikdeMelo2020ExplainableLinkPrediction,WuEtAl2019EfficientlyEmbeddingDynamic,DarunaEtAl2021ContinualLearningKnowledge}.
To highlight this distinction, we use the term \emph{evolving knowledge graph} to refer to knowledge graphs that change over time, and the term \emph{static knowledge graph} to refer to those that do not.

Currently, there are practically no published datasets recording a knowledge graph's organic evolution over time that would enable such research and facilitate reproducible evaluation environments.
Instead, existing research either simulates knowledge graph evolution on top of datasets of static knowledge graphs using simple heuristics such as statement ordering (e.\,g., Daruna et al.~\cite{DarunaEtAl2021ContinualLearningKnowledge} split static knowledge graph datasets into chunks based on the order of triples in the dataset) or are based on the computed changes between major releases of knowledge graphs (e.\,g., Wu et al.~\cite{WuEtAl2019EfficientlyEmbeddingDynamic} calculate the change sets between YAGO2.5, YAGO3, and YAGO3.1~\cite{RebeleEtAl2016YAGOMultilingualKnowledge}).
While the former approaches can hardly be argued to constitute evolving knowledge graphs---in particular statement updates or deletion are not modeled---the latter ones fail to capture the inherent dynamics of how changes occur on the individual level, for example that popular entities receive frequent updates or that some updates might be (partially) reversed after a few days.

Wikidata~\cite{VrandecicKrotzsch2014WikidataFreeCollaborative} is \enquote{a collaboratively edited knowledge-base \textelp{} whose aim is to curate and represent the factual information of Wikipedia (across all languages) in an interoperable, machine-readable format}~\cite{HernandezEtAl2015ReifyingRDFWhat}.
With 90 million entities and 1.4 billion revision made by 20 thousand active users\footnote{Statistics from \url{https://www.wikidata.org/wiki/Wikidata:Statistics} (23~July~2021).}, Wikidata is the prime example of an evolving knowledge graph.
In this paper, we present \wikidated, an evolving knowledge graph dataset covering the full revision history of Wikidata.
To the best of our knowledge, \wikidated\ is the first large dataset of an evolving knowledge graph.
It records the fine-grained, organic evolution of Wikidata since its inception in 2012 until June 2021.
It is suited for research into how knowledge graphs and their communities change over time---specifically for Wikidata, such as done in Sarasua et al.~\cite{SarasuaEtAl2019EvolutionPowerStandard}---and enables reproducible evaluation environments for indexing and representation approaches of evolving knowledge graphs, such as incremental knowledge graph embedding~\cite{WuEtAl2019EfficientlyEmbeddingDynamic,DarunaEtAl2021ContinualLearningKnowledge}.

In particular, our contributions are:
\begin{itemize}
    \item We present our methodology for transforming a dump of Wikidata's revision history into streams of RDF triple deletions and additions which form \wikidated, a dataset recording Wikidata's evolution over time~(\cref{sec:construction}).
    \item We present statistics over the dataset and visualize its characteristics~(\cref{sec:stat}).
    \item We publicly release \wikidated\ in two variants, the code base used to built it, and a Python API to access it\cref{note:download}.
\end{itemize}
In addition to the above, \cref{sec:background,sec:related} discuss background and related work, respectively, and \cref{sec:concl} concludes.
    \section{Background}
\label{sec:background}

In this section, we review the data model of Wikidata~(\cref{sec:data-model}) and its serialization as RDF~(\cref{sec:rdf}).

\subsection{Data Model}
\label{sec:data-model}

Formally, the data model of Wikidata\footnote{For more details, see \url{https://www.mediawiki.org/wiki/Wikibase/DataModel}.} can be defined as a set of \emph{entities} $e_1, \dots, e_N$, where $N$ is the number of entities in Wikidata.
Let $\mathrm{id}(e_i)$ denote the \emph{entity ID} of entity $e_i$.
Each entity is either an \emph{item} or a \emph{property}\footnote{As discussed in \cref{sec:limits}, technically, entities can also be \emph{lexemes}, \emph{forms}, or \emph{senses}.}.
Items are things or concepts in the real world about which facts should be stored; their IDs are numbers prefixed with~\enquote{Q}---for example, the English writer \textsf{Douglas Adams}~(Q42) or the concept of a \textsf{human}~(Q5).
Properties are abstract types of statements which are used to store facts about entities; their IDs are numbers prefixed with~\enquote{P}---for example, the properties \textsf{instance-of}~(P31) or \textsf{date-of-birth}~(P569).

A \emph{revision} defines an entity's state at a specific point in time.
Each revision is comprised of: (1) a \emph{fingerprint}, which consists of multilingual sets of \emph{labels}, \emph{descriptions}, and \emph{aliases} of the entity, (2) a set of \emph{site links}, which are usually links to Wikipedia articles about the entity, and (3) a set of \emph{statements}, which are records of facts about the entity.
There is one exception: if an entity is found to be a duplicate of another one, a revision can also be a \emph{redirect}.
In the case of redirects, no fingerprint, site links, or statements are present, and the revision consists of just the entity ID that the redirect's entity is deemed to be a duplicate of.
Further, each revision (including redirects) carries metadata, such as the time it was created at, the contributor that authored it, and a comment string about the change.

Every time an entity is modified, a new revision is created.
We therefore model entities as sequences of their revisions $e_i = (r_{i, 1}, \dots, r_{i, n_i})$, where $n_i$ is the number of revisions of entity $e_i$.
Let $\mathrm{id}(r_{i, j})$ denote the \emph{revision ID} of revision $r_{i, j}$.
Revision IDs are assigned by incrementing a global counter.
They are thereby unique over all entities and induce a total ordering of all revisions in Wikidata\footnote{However, the special case, in which a revision has an earlier timestamp than one with a lower ID, can occur. We attribute this to slight miss-synchronizations of clocks on parallel servers. The difference is never larger than one second.}.

Finally, statements record facts about entities and fundamentally consist of a property and a \emph{value}, which is either another entity or a literal.
For example, statements about \textsf{Douglas Adams} include \textsf{instance-of human} and \textsf{date-of-birth \enquote{11~March~1952}}.
As is the case in the latter example, literals can be of various data types, e.\,g., dates or geographical coordinates.
An example of a statement about a property is that the \textsf{complementary-property} of \textsf{date-of-birth} is \textsf{date-of-death}.
Note, that entities whose latest revision is a redirect and deleted entities can still be targets of the statements of other entities, but that Wikidata aims to replace instances of this with new revisions where this is not the case.
There are two special values of statements: \textsf{none}, which signifies that it is known that the property of that entity has no value, and \textsf{some}, which indicates that it is known that there is some value for the property of that entity, but it is unknown what it is.
Each statement can be annotated by (1) a set of \emph{qualifiers}, which refine a statement (e.\,g., to indicate that it has only been true for some period of time), (2) a set of \emph{references}, which provide sources to support the statement, and (3) a \emph{rank}, which can be used to assign preference to conflicting statements (e.\,g., to distinguish current from historical facts).

Dumps of Wikidata's contents are available for download in various formats\footnote{See \url{https://www.wikidata.org/wiki/Wikidata:Database_download}.}.
Most dump formats only provide the most recent revision of each entity; only the \textsf{pages-meta-history} XML dumps include the full revision history, but store revision contents as JSON blobs\footnote{See \url{https://doc.wikimedia.org/Wikibase/master/php/md_docs_topics_json.html}.}.
Notably, each revision stores its complete state at its creation time and there is no trivial way to identify what changed from one revision to the next.

In cases of vandalism, or when entities do not meet Wikidata's notability policy\footnote{Available at \url{https://www.wikidata.org/wiki/Wikidata:Notability}.}, administrators may delete the affected revisions or whole entities from Wikidata\footnote{Requests for deletions and the decisions for each are recorded at \url{https://www.wikidata.org/wiki/Wikidata:Requests_for_deletions}.}.
Since such deleted contents may contain copyrighted materials or sensible personal information, deleted entities and revisions are not accessible to the general public and are not contained in the official dumps of Wikidata.
IDs of deleted entities and revisions are never reused for new ones, which leads to gaps in the incremental numbering.
In case an entity is detected that is a duplicate of an existing one, Wikidata prefers not to delete the new entity, but to establish a redirect from the new entity to the existing one instead.

\subsection{RDF}
\label{sec:rdf}

The \emph{Resource Description Framework~(RDF)}~\cite{SchreiberRaimond2014RDFPrimer,CyganiakEtAl2014RDFConceptsAbstract} is a metadata format and the standard way of exchanging information on the Semantic Web.
For the \wikidated\ dataset, we serialize Wikidata revisions as RDF graphs which allows for a straightforward definition of change between revisions.

Let $I$, $B$, and $L$ be disjoint countably infinite sets of IRIs, blank nodes, and literals, respectively.
A \emph{RDF triple} is a triple $(s, p, o) \in (I \cup B) \times I \times (I \cup B \cup L)$, where $s$ is called the subject, $p$ the predicate, and $o$ the object.
A \emph{RDF graph} is a set of RDF triples.
Let $G_1 \setminus G_2$ denote the \emph{set difference} between two RDF graphs $G_1$ and $G_2$.
Computing it is non-trivial, because in order to decide whether two RDF triples are equal, one needs to decide which triple components are equal to one another.
This is straightforward for IRIs and literals, but hard for blank nodes, as they are only characterized through the RDF triples they participate in and do not have identifiers across RDF graphs.
In general, finding a mapping between the blank nodes of two RDF graphs that minimizes the set difference is NP-hard~\cite{TummarelloEtAl2007RDFSyncEfficientRemote,PapavasileiouEtAl2013HighlevelChangeDetection,AhnEtAl2014GDiffGroupingAlgorithm,LantzakiEtAl2017RadiusawareApproximateBlank}.
For \wikidated, we circumvent this issue~(see \cref{sec:impl}).

\begin{lstlisting}[language=rdf, caption={RDF serialization of Wikidata entity Q42 in Turtle syntax (abridged).}, float, label=lst:q42rdf, escapechar=\#]
wd:Q42 a wikibase:Item ;
    rdfs:label "Douglas Adams"@en ;#\label{line:q42rdf-label}#
    schema:description "English writer and humorist"@en ;#\label{line:q42rdf-desc}#
    skos:altLabel "Douglas Noel Adams"@en ;#\label{line:q42rdf-alias}#
    wdt:P569 "1952-03-11T00:00:00Z"^^xsd:dateTime ;#\label{line:q42rdf-simple}#
    p:P569 s:Q42-D8404CDA-25E4-4334-AF13-A3290BCD9C0F .#\label{line:q42rdf-full-decl}#
    
<https://en.wikipedia.org/wiki/Douglas_Adams> a schema:Article ;#\label{line:q42rdf-sitelink-start}#
    schema:about wd:Q42 .#\label{line:q42rdf-sitelink-end}#
    
s:Q42-D8404CDA-25E4-4334-AF13-A3290BCD9C0F a wikibase:Statement ;#\label{line:q42rdf-full-start}#
    ps:P569 "1952-03-11T00:00:00Z"^^xsd:dateTime ;
    prov:wasDerivedFrom ref:355b56329b78db22be549dec34f2570ca61ca056 .#\label{line:q42rdf-full-end}\label{line:q42rdf-ref-decl}#
    
ref:355b56329b78db22be549dec34f2570ca61ca056 a wikibase:Reference ;#\label{line:q42rdf-ref-start}#
    pr:P248 wd:Q5375741 .#\label{line:q42rdf-ref-end}#
\end{lstlisting}

While Wikidata doesn't store revisions in RDF internally, RDF serializations for them are available~\cite{ErxlebenEtAl2014IntroducingWikidataLinked,HernandezEtAl2015ReifyingRDFWhat,MalyshevEtAl2018GettingMostOut}.
\Cref{lst:q42rdf} shows an example\footnote{Taken from \url{https://www.wikidata.org/wiki/Special:EntityData/Q42.ttl}. Documentation at \url{https://www.mediawiki.org/wiki/Wikibase/Indexing/RDF_Dump_Format}.}.
The shown revision of entity Q42 (\textsf{Douglas Adams}) is described by its fingerprint (\cref{line:q42rdf-label,line:q42rdf-desc,line:q42rdf-alias}), a site link (\crefrange{line:q42rdf-sitelink-start}{line:q42rdf-sitelink-end}), a \emph{simple statement} (\cref{line:q42rdf-simple}), and a \emph{full statement} (\cref{line:q42rdf-full-decl} and \crefrange{line:q42rdf-full-start}{line:q42rdf-full-end}).
Both statements specify a \textsf{date-of-birth} (P569) of \enquote{11 March 1952}.
The difference is that simple statements give a value \enquote{directly} while discarding statement annotations (i.\,e., qualifiers, references, and ranks), whereas full statements use \emph{reification} (the insertion of a special statement node) to facilitate annotations.
In this case, a reference to \textsf{Encyclopædia Britannica Online} (Q5375741, \crefrange{line:q42rdf-ref-start}{line:q42rdf-ref-end}) is used to annotate the statement (\cref{line:q42rdf-ref-decl}).
    \section{Constructing \wikidated\ from Wikidata Dumps}
\label{sec:construction}

In this section, we discuss our methodology for creating \wikidated~(\cref{sec:method}), our implementation~(\cref{sec:impl}), and limitations~(\cref{sec:limits}).

\subsection{Methodology}
\label{sec:method}

Fundamentally, \wikidated\ is a transformation from a Wikidata dump's stream of revisions into a stream of \emph{incremental revisions}.
We define an incremental revision as a tuple of the (1)~entity metadata (the entity ID and some Wikidata internal fields), and the (2)~revision metadata (the revision ID and when and by whom it was authored) of the Wikidata revision it is based upon, as well as sets of (3)~\emph{RDF triple deletions} and (4)~\emph{RDF triple additions} in comparison to the previous revision of the respective entity.
As hinted at in \cref{sec:background}, we thus decide to define change between Wikidata revisions via the difference in triples between their RDF serializations.
This allows for more straightforward dataset modeling and consumption, as opposed to defining change for each of the different aspects of the Wikidata data model~(fingerprint, site links, and statements with qualifiers, references, and ranks).

\wikidated\ consists of two complementary variants of the same data:
\begin{enumerate}
    \item The \emph{global-stream} variant consists of all incremental revisions across all entities sorted in chronological order.
    \item The \emph{entity-streams} variant contains a separate stream of incremental revisions for each entity of Wikidata.
\end{enumerate}
The former can be used for global analysis, e.\,g., for analyzing the number or style of revisions in a specific time period, whereas the latter is useful for entity-centered analysis, e.\,g., when one is only interested in a subset of all entities or when the aim is to directly compare consecutive revisions of the same entity.

\begin{algorithm}[tb]
    $\Delta_\textsf{Global} \gets$ empty sequence\;\label{line:wikidated-init-global}
    \BlankLine\BlankLine
    download pages-meta-history dump of Wikidata\;\label{line:wikidated-dl}
    \ForEach{entity $e_i$ \KwSty{in} dump}{\label{line:wikidated-iter-entity}
        $\Delta_\textsf{Entity} \gets$ empty sequence\;\label{line:wikidated-init-entity}
        \BlankLine\BlankLine
        $r_\textsf{RDF-prev} \gets \{\}$\;\label{line:wikidated-rdf-prev}
        \For{$j = 1$ \KwTo $n_i$}{\label{line:wikidated-iter-revs}
            $r_\textsf{Meta} \gets$ take revision metadata of $r_{i, j}$ from dump\;\label{line:wikidated-meta}
            $r_\textsf{JSON} \gets$ take JSON blob of $r_{i, j}$ from dump\;\label{line:wikidated-json}
            \BlankLine\BlankLine
            $r_\textsf{RDF} \gets$ serialize $r_\textsf{JSON}$ as RDF graph\;\label{line:wikidated-rdf}
            $r_\textsf{Del} \gets r_\textsf{RDF-prev} \setminus r_\textsf{RDF}$\;\label{line:wikidated-del}
            $r_\textsf{Add} \gets r_\textsf{RDF} \setminus r_\textsf{RDF-prev}$\;\label{line:wikidated-add}
            $r_\textsf{RDF-prev} \gets r_\textsf{RDF}$\;\label{line:wikidated-rdf-prev-update}
            \BlankLine\BlankLine
            append incremental revision $(e_i, r_\textsf{Meta}, r_\textsf{Del}, r_\textsf{Add})$ to $\Delta_\textsf{Entity}$\;\label{line:wikidated-append-entity}
        }
        \BlankLine
        \KwSty{output} $\Delta_\textsf{Entity}$ as \emph{entity-stream} variant of entity $e_i$\;\label{line:wikidated-export-entity}
        append all elements of $\Delta_\textsf{Entity}$ to $\Delta_\textsf{Global}$\;\label{line:wikidated-append-global}
    }
    \BlankLine
    sort $\Delta_\textsf{Global}$ after ascending revision IDs $\mathrm{id}(r_{i, j})$ across entities\;\label{line:wikidated-sort-global}
    \KwSty{output} $\Delta_\textsf{Global}$ as \emph{global-stream} variant\;\label{line:wikidated-export-global}
    \caption{Construction of the \wikidated\ dataset.}
    \label{alg:wikidated}
\end{algorithm}

\Cref{alg:wikidated} outlines the steps of creating \wikidated.
First, we download a full Wikidata dump\footnote{Specifically, \wikidated\ is based on the \textsf{20210601-pages-meta-history} dump, the history of Wikidata from its inception on 30~October~2012 until June~2021. The ID of the last revision is 1433475551, which was authored on 2 June 2021 at 05:35:58.}~(\cref{line:wikidated-dl}).
Next, we iterate over all entities in the dump~(\cref{line:wikidated-iter-entity}).
For each entity, we iterate over all of its revisions in the order they were created in~(\cref{line:wikidated-iter-revs}).
Because the dump files store entities and revisions in exactly this order, this amounts to linearly traversing the dump files.
For each revision, we first extract revision metadata and the JSON blob of revision contents from the dump files~(\cref{line:wikidated-meta,line:wikidated-json}).
We then serialize the revision contents as an RDF graph~(\cref{line:wikidated-rdf}), and compute the sets of RDF triple deletions and additions~(\cref{line:wikidated-del,line:wikidated-add}) compared to the RDF graph of the previous revision, for which we maintain a helper variable~(\cref{line:wikidated-rdf-prev,line:wikidated-rdf-prev-update}).
Having now transformed the revision into its incremental counterpart, we append it to a sequence of all incremental revisions of the entity $\Delta_\textsf{Entity}$~(\cref{line:wikidated-append-entity}, initialization in \cref{line:wikidated-init-entity}).
After all revisions of an entity have been iterated over, we output~$\Delta_\textsf{Entity}$ as the entity-stream variant of that entity~(\cref{line:wikidated-export-entity}).
Finally, we maintain a sequence of all incremental revisions across all entities~$\Delta_\textsf{Global}$~(\cref{line:wikidated-init-global,line:wikidated-append-global}).
After sorting all revisions in it globally~(\cref{line:wikidated-sort-global}), we also output $\Delta_\textsf{Global}$ as the global-stream variant~(\cref{line:wikidated-export-global}).

\subsection{Implementation}
\label{sec:impl}

We provide a Python API for browsing and iterating the dataset without having to know how the dataset is stored on disk.
Internally, both variants are stored as \texttt{gzip}-compressed text files in JSON Lines format, i.\,e., each line is a JSON object encoding one incremental revision.
For the entity-streams variant, the files for all entities are packaged in a \texttt{tar} archive.
While approaches for storing differences between RDF graphs in RDF itself exist~\cite{Berners-LeeConnolly2004DeltaOntologyDistribution,PellissierTanonSuchanek2019QueryingEditHistory}, we hereby opt for a less Semantic-Web-oriented distribution format, because we feel that it allows for easier consumption by most users.
We are open to releasing the dataset in RDF later (based on community demand).

The file size of the global-stream variant of \wikidated\ is 239\,GiB whereas the \texttt{tar} archive for the entity-streams variant is 329\,GiB (both \texttt{gzip}-compressed).
In contrast to this, the official Wikidata dump of non-incremental revisions that \wikidated\ is built from is available in the two compression formats \texttt{bz2} and \texttt{7z} with a size of 1\,040\,GiB and 339\,GiB, respectively.
While \wikidated's smallest variant is thus only 71\% the size of the official dump's smallest format, one might have expected an even larger reduction in size due to the usage of incremental revisions that do not repeat all statements from earlier revisions.
We suspect that this benefit is offset by using the RDF serialization, which is more verbose than the JSON blobs of the official dumps, that the non-incremental revisions are more compressible through their repeated statements, and by the inferior compression of \texttt{gzip} compared to \texttt{7z}.
We are therefore looking into publishing our dataset in additional compression formats in the future, but have opted for the universally-available and streamable \texttt{gzip} format for the first release.

For serializing Wikidata revisions as RDF graphs~(\cref{line:wikidated-rdf}), we use Wikidata Toolkit\footnote{Available at \url{https://www.mediawiki.org/wiki/Wikidata_Toolkit}.}.
Because it does not provide a way to serialize Wikidata revisions that are redirects, we encode these via the \lstinline{owl:sameAs} predicate, following the choice of the Wikidata Query Service\footnote{Available at \url{https://query.wikidata.org/}.}~\cite{MalyshevEtAl2018GettingMostOut}.
Additionally, we discard all RDF triples from Wikidata Toolkit's output that are not directly related to the entity at hand.
The triples discarded in this manner contain ontological information about Wikidata concepts such as items and properties.
If needed, these can always be reconstructed from context and they never change between revisions.

For computing set differences between RDF graphs~(\cref{line:wikidated-del,line:wikidated-add}), we can avoid the difficult task of finding an optimal mapping between blank nodes.
In the case of Wikidata's RDF serialization, blank nodes are only used to encode the special \textsf{some} values, and each blank node never occurs in more than one RDF triple.
Because of this, there is an efficient way to determine the set difference between RDF serializations of two Wikidata revisions: two RDF triples can be treated as equal if their non-blank-node components are equal.

Last, assembling the incremental revision streams $\Delta_\textsf{Global}$ and $\Delta_\textsf{Entity}$~(\cref{line:wikidated-append-global,line:wikidated-sort-global}) is not as straightforward as presented, because both would quickly exceed available main memory.
Instead, we directly append any incremental revisions~(\cref{line:wikidated-append-entity}) to the target file without keeping them in memory.
To merge all $\Delta_\textsf{Entity}$ streams into a sorted $\Delta_\textsf{Global}$, we use a hierarchical multiway merge.

\subsection{Limitations}
\label{sec:limits}

The main limitation of \wikidated\ is that is does not contain any record of deleted entities or revisions\footnote{In \cref{sec:related}, we review the work of Shenoy et al.~\cite{ShenoyEtAl2021StudyQualityWikidata}, which describes an approach that is able to obtain some information about deleted entities (from monthly dumps).} because these are not recorded in Wikidata's revision history dumps, as explained in \cref{sec:data-model}.
However, entity deletions are comparatively rare in Wikidata, since the common case of duplicate entities is addressed through merges, i.\,e., redirects of one entity to another, which---unlike deletions---are recorded in \wikidated.
Additionally, statements of other entities may still target deleted entities, so partial history of them is recorded.

Multiple implementations of RDF serializations of Wikidata revisions exist.
The one in Wikidata Toolkit was used to construct \wikidated.
Both the RDF exports of individual Wikidata entities and the official Wikidata RDF dumps use two other slightly different implementations.
In practice, the differences are minimal and mostly amount to how certain metadata is encoded---the important parts, i.\,e., facts about entities, are encoded identically in all implementations.

By design, \wikidated\ only contains the revision history of entities.
As a consequence the \enquote{meta level} of Wikidata is not part of the dataset.
Among other things, this includes the talk pages of all entities, where editors discuss aspects such as how certain content should be modeled or what is in scope for Wikidata, or the help pages, which document how to use Wikidata.
While this plain text data is part of the Wikidata dumps it is not RDF serializable.

In May 2018, three new entity types have been added to the Wikidata data model to model lexicographical data: lexemes, forms, and senses\footnote{Documentation at \url{https://www.wikidata.org/wiki/Wikidata:Lexicographical_data}.}~\cite{Nielsen2020LexemesWikidata2020}.
While these are part of the Wikidata dumps and a RDF serialization for them has been defined~\cite{Nielsen2020LexemesWikidata2020}, it has not yet been implemented in Wikidata Toolkit and lexicographical data is thus not part of this first release of our dataset.
    \section{\wikidated\ Dataset Characteristics}
\label{sec:stat}

In this section, we present statistical characteristics of the \wikidated\ dataset in order to provide context for any research work building upon it.
Much of the analysis also applies to Wikidata given that \wikidated\ is a direct representation of it.

\begin{figure}[tb]
    \stepcounter{figure}
    \begin{minipage}[b]{0.4\textwidth}
        \centering
        \refstepcounter{subfigure}\label{fig:num-entities-revisions-over-time}
        \definecolor{colorEntities}{HTML}{E7298A}
\definecolor{colorRevisions}{HTML}{7570B3}
\ifstandalone
    \begin{tikzpicture}
\else
    \begin{tikzpicture}[baseline, trim axis left, trim axis right]
\fi
    \begin{axis}[
        scale only axis,
        set layers,
        width=\ifstandalone0.4\fi\textwidth,
        height=2.75cm,
        date coordinates in=x,
        enlargelimits=0.05,
        xmin=2012-01-01,
        xmax=2020-01-01,
        axis x line=none,
        ymin=0,
        ymax=1.5e9,
        axis y line*=right,
        ylabel={Number of revisions\\in billions},
        ylabel style={align=center},
        ytick={0,3e8,6e8,9e8,1.2e9,1.5e9},
        yticklabels={0,0.3,0.6,0.9,1.2,1.5},
        major y tick style={draw=none},
        minor y tick num=2,
        scaled y ticks=false,
    ]
        \addplot[colorRevisions, densely dashed, thick, mark=*, mark options={solid, fill=colorRevisions!50}] coordinates {
            (2012-01-01,2828229)
            (2013-01-01,94681367)
            (2014-01-01,180330581)
            (2015-01-01,280766284)
            (2016-01-01,413809442)
            (2017-01-01,603637286)
            (2018-01-01,810294194)
            (2019-01-01,1070094680)
            (2020-01-01,1311636649)
        };
        \label{plot_revisions}
    \end{axis}
    \begin{axis}[
        scale only axis,
        set layers,
        width=\ifstandalone0.4\fi\textwidth,
        height=2.75cm,
        date coordinates in=x,
        enlargelimits=0.05,
        xmin=2012-01-01,
        xmax=2020-01-01,
        xtick={2012-01-01,2013-01-01,2014-01-01,2015-01-01,2016-01-01,2017-01-01,2018-01-01,2019-01-01,2020-01-01},
        xticklabels={2012\vphantom{$\geq$},,2014\vphantom{$\geq$},,2016\vphantom{$\geq$},,2018\vphantom{$\geq$},,2020\vphantom{$\geq$}},
        xtick pos=bottom,
        xtick align=outside,
        xlabel={End of year\vphantom{p}},
        grid=major,
        ymin=0,
        ymax=1e8,
        axis y line*=left,
        ylabel={Number of entities\\in millions},
        ylabel style={align=center},
        ylabel shift=-0.05cm,
        ytick={0,2e7,4e7,6e7,8e7,1e8},
        yticklabels={0,20,40,60,80,100},
        major y tick style={draw=none},
        minor y tick num=3,
        scaled y ticks=false,
        legend cell align={left},
        legend style={at={(0.08,0.96)}, anchor=north west},
    ]
        \addplot[colorEntities, solid, thick, mark=triangle*, mark options={fill=colorEntities!50}] coordinates {
            (2012-01-01,2031106)
            (2013-01-01,13990804)
            (2014-01-01,16768626)
            (2015-01-01,19784537)
            (2016-01-01,25433838)
            (2017-01-01,43530247)
            (2018-01-01,55402670)
            (2019-01-01,74731553)
            (2020-01-01,94543677)
        };
        \addlegendentry{Entities}
        \addlegendimage{/pgfplots/refstyle=plot_revisions}
        \addlegendentry{Revisions}
    \end{axis}
\end{tikzpicture}
\iffalse\end{tikzpicture}\fi % Workaround to not confuse Overleaf's syntax highlight.
        \\[0.05cm]\textbf{(a)}
        % Total number of entities: 96 646 606
        % Total number of revisions: 1 411 008 075
    \end{minipage}
    \hfill
    \begin{minipage}[b]{0.4\textwidth}
        \centering
        \refstepcounter{subfigure}\label{fig:num-revisions-per-entity}
        \definecolor{colorRevisions}{HTML}{7570B3}
\ifstandalone
    \begin{tikzpicture}
\else
    \begin{tikzpicture}[baseline, trim axis left]
\fi
    \begin{axis}[
        ybar,
        scale only axis,
        width=\ifstandalone0.4\fi\textwidth,
        height=2.75cm,
        bar width=0.9cm,
        point meta={y*100},
        enlarge x limits=0.2,
        xlabel={Number of revisions per entity},
        xtick={0,...,4},
        xticklabels={
            1\vphantom{$\geq$},
            2--9\vphantom{$\geq$},
            10--99\vphantom{$\geq$},
            $\geq$\,100,
        },
        xtick pos=bottom,
        xtick align=outside,
        enlarge y limits={upper, value=0.2},
        ylabel={Relative frequency},
        ylabel near ticks,
        yticklabel pos=right,
        ytick style={draw=none},
        ymin=0,
        yticklabel={\pgfmathparse{\tick*100}\pgfmathprintnumber{\pgfmathresult}\%},
        ymajorgrids,
        nodes near coords={\pgfmathprintnumber\pgfplotspointmeta\%},
        nodes near coords style={font=\footnotesize},
    ]
        \addplot[thick, draw=colorRevisions, fill=colorRevisions!33] coordinates {
            (0,0.025)
            (1,0.572)
            (2,0.394)
            (3,0.010)
        };
    \end{axis}
\end{tikzpicture}
\iffalse\end{tikzpicture}\fi % Workaround to not confuse Overleaf's syntax highlight.
        \\[0.05cm]\textbf{(b)}
        % mean: 14.5997
        % std: 24.9169
        % std_mean: 0.0025
        % 1%-quantile: 1
        % 2.5%-quantile: 2
        % 25%-quantile: 4
        % median: 7
        % 75%-quantile: 16
        % 97.5%-quantile: 68
        % 99%-quantile: 99
        % welford mean: 14.599665041522524
        % welford std: 24.91687113450437
    \end{minipage}
    \addtocounter{figure}{-1}
    \caption{
        \textbf{(a)}~Number of entities and revisions at the end of each year.
        The totals by 2~June~2021 (end of dataset's time range) are 96.6 million entities and 1.4 billion revisions, respectively.
        \textbf{(b)}~Histogram of number of revisions per entity (mean 14.60, standard deviation 24.94, median 7).}
\end{figure}
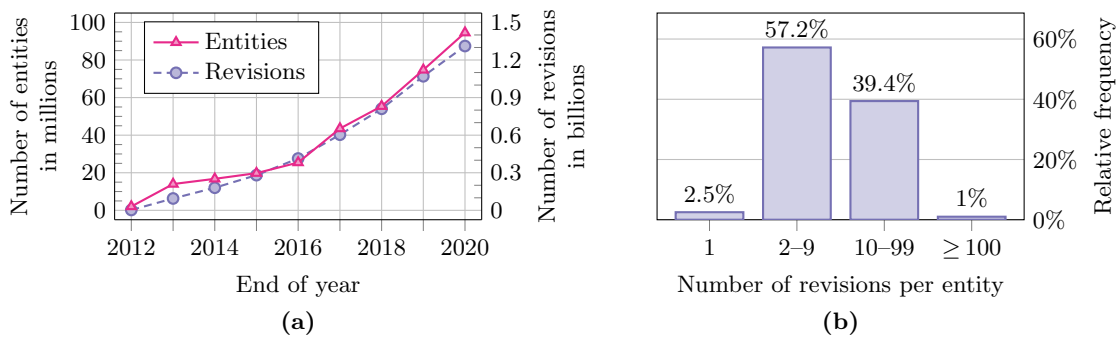

We plot the number of entities and revisions over time in \cref{fig:num-entities-revisions-over-time}\footnote{In all figures of this section, we treat revisions that are redirects as regular revisions (being represented by exactly one \lstinline{owl:sameAs} RDF triple). As a consequence, we also count entities whose latest revision is a redirect as regular entities.}.
For both, we observe a mostly linear growth, with a slightly stronger incline since 2016, which follows the integration of Freebase into Wikidata in the latter half of 2015~\cite{TanonEtAl2016FreebaseWikidataGreat}.
Interestingly, this means that the ratio between both frequencies is roughly constant at about 14 revisions per entity.

\cref{fig:num-revisions-per-entity} shows the number of revisions per entity in more detail.
Roughly half of all entities have fewer than 10 revisions, and the majority have more than one.
99\% of entities have less than 100 revisions, whereas some of the remaining entities can have significantly more.
\idea{Scatter plot comparing age of entity to number of revisions?}

\begin{figure}[tb]
    \stepcounter{figure}
    \begin{minipage}[b]{0.55\textwidth}
        \centering
        \refstepcounter{subfigure}\label{fig:time-between-revisions}
        \definecolor{colorRevisions}{HTML}{7570B3}
\ifstandalone
    \begin{tikzpicture}
\else
    \begin{tikzpicture}[baseline, trim axis left]
\fi
    \begin{axis}[
        scale only axis,
        width=\ifstandalone0.55\fi\textwidth,
        height=2.75cm,
        bar width=0.9cm,
        enlarge x limits=0.04,
        xlabel={Time between consecutive revs. of same entity\vphantom{/}},
        xtick={0,...,8},
        xticklabels={
            0,
            1\,s,
            1\,min,
            1\,h,
            1\,d,
            7\,d,
            30\,d,
            1\,yr,
            $\infty$\vphantom{1y},
        },
        xtick pos=bottom,
        xtick align=outside,
        enlarge y limits={upper, value=0.2},
        ylabel={Relative frequency},
        ymin=0,
        yticklabel={\pgfmathparse{\tick*100}\pgfmathprintnumber{\pgfmathresult}\%},
        ytick style={draw=none},
        ymajorgrids,
    ]
        \addplot[ybar interval, thick, draw=colorRevisions, fill=colorRevisions!33] coordinates {
            (0,0.1678)
            (1,0.1315)
            (2,0.0313)
            (3,0.0546)
            (4,0.0751)
            (5,0.1470)
            (6,0.3535)
            (7,0.0393)
            (8,0.0393)
        };
        \draw[anchor=south, font=\footnotesize]
            (axis cs: 0.5,0.1678) node {16.8\%\ }
            (axis cs: 1.5,0.1315) node {\ 13.2\%}
            (axis cs: 2.5,0.0313) node {3.1\%}
            (axis cs: 3.5,0.0546) node {5.5\%}
            (axis cs: 4.5,0.0751) node {7.5\%}
            (axis cs: 5.5,0.1470) node {14.7\%\ \ }
            (axis cs: 6.5,0.3535) node {35.3\%}
            (axis cs: 7.5,0.0393) node {3.9\%}
        ;
    \end{axis}
\end{tikzpicture}
\iffalse\end{tikzpicture}\fi % Workaround to not confuse Overleaf's syntax highlight.
        \\[0.05cm]\textbf{(a)}
        % mean: 70.6566
        % std: 149.3065
        % std_mean: 0.0041
        % 1%-quantile: 1
        % 2.5%-quantile: 1
        % 25%-quantile: 1
        % median: 12
        % 75%-quantile: 79
        % 97.5%-quantile: 452
        % 99%-quantile: 655
        % welford mean: 6046514.6788
        % welford std: 12907079.9985
    \end{minipage}
    \hfill
    \begin{minipage}[b]{0.4\textwidth}
        \centering
        \refstepcounter{subfigure}\label{fig:num-triple-adds-dels-per-revision}
        \definecolor{colorDeletions}{HTML}{D95F02}
\definecolor{colorAdditions}{HTML}{1B9E77}
\ifstandalone
    \begin{tikzpicture}
\else
    \begin{tikzpicture}[baseline, trim axis left]
\fi
    \begin{axis}[
        ybar,
        scale only axis,
        width=\ifstandalone0.4\fi\textwidth,
        height=2.75cm,
        bar width=0.45cm,
        point meta={y*100},
        enlarge x limits=0.2,
        xlabel={Triple additions/deletions per rev.},
        xtick={0,...,3},
        xticklabels={
            0\vphantom{y},
            1\vphantom{y},
            2--9\vphantom{y},
            $\geq$\,10\vphantom{y}
        },
        xtick pos=bottom,
        enlarge y limits={upper, value=0.2},
        ylabel={Relative frequency},
        ylabel near ticks,
        yticklabel pos=right,
        ymin=0,
        yticklabel={\pgfmathparse{\tick*100}\pgfmathprintnumber{\pgfmathresult}\%},
        ytick style={draw=none},
        nodes near coords={\pgfmathprintnumber\pgfplotspointmeta\%},
        nodes near coords style={font=\footnotesize},
        ymajorgrids,
        legend entries={Additions, Deletions},
        legend pos=north east,
        legend cell align={left},
    ]
        \addplot[thick, draw=colorAdditions, preaction={fill=colorAdditions!33}, pattern=crosshatch dots, pattern color=colorAdditions] coordinates {
            (0,0.04)
            (1,0.46)
            (2,0.39)
            (3,0.11)
        };
        \addplot[thick, draw=colorDeletions, fill=colorDeletions!32] coordinates {
            (0,0.80)
            (1,0.11)
            (2,0.08)
            (3,0.01)
        };
    \end{axis}
\end{tikzpicture}
\iffalse\end{tikzpicture}\fi % Workaround to not confuse Overleaf's syntax highlight.
        \\[0.05cm]\textbf{(b)}
        % statistics (additions)
        %   mean: 9.5630
        %   std: 38.6020
        %   std_mean: 0.0010
        %   1%-quantile: 0
        %   2.5%-quantile: 0
        %   25%-quantile: 1
        %   median: 2
        %   75%-quantile: 6
        %   97.5%-quantile: 81
        %   99%-quantile: 117
        %   welford mean: 9.5629
        %   welford std: 38.6019
        % statistics (deletions)
        %   mean: 0.8828
        %   std: 14.7316
        %   std_mean: 0.0004
        %   1%-quantile: 0
        %   2.5%-quantile: 0
        %   25%-quantile: 0
        %   median: 0
        %   75%-quantile: 0
        %   97.5%-quantile: 7
        %   99%-quantile: 13
        %   welford mean: 0.8827
        %   welford std: 14.7316
    \end{minipage}
    \vspace{0.025\textwidth}
    
    \begin{minipage}[b]{0.55\textwidth}
        \centering
        \refstepcounter{subfigure}\label{fig:time-until-triple-add-del}
        \definecolor{colorDeletions}{HTML}{D95F02}
\definecolor{colorAdditions}{HTML}{1B9E77}
\ifstandalone
    \begin{tikzpicture}
\else
    \begin{tikzpicture}[baseline, trim axis left]
\fi
    \begin{axis}[
        ybar,
        scale only axis,
        width=\ifstandalone0.55\fi\textwidth,
        height=2.75cm,
        bar width=0.5cm,
        point meta={y*100},
        enlarge x limits=0.15,
        xlabel={Time until triple is first added/deleted in days},
        xtick={0,...,4},
        xticklabels={{$[0,1]$},{$(\cdot,30]$},{$(\cdot,180]$},{$(\cdot,365]$},{$(\cdot,\infty)$}},
        xtick pos=bottom,
        enlarge y limits={upper, value=0.2},
        ylabel={Relative frequency},
        ymin=0,
        yticklabel={\pgfmathparse{\tick*100}\pgfmathprintnumber{\pgfmathresult}\%},
        ytick style={draw=none},
        max space between ticks=15,
        ymajorgrids,
        nodes near coords={\pgfmathprintnumber\pgfplotspointmeta\%},
        nodes near coords style={font=\footnotesize},
        legend entries={First addition, First deletion},
        legend style={at={(0.5,0.97)}, anchor=north},
        legend cell align={left},
    ]
        \addplot[thick, draw=colorAdditions, preaction={fill=colorAdditions!33}, pattern=crosshatch dots, pattern color=colorAdditions] coordinates {
            (0,0.48)
            (1,0.07)
            (2,0.08)
            (3,0.09)
            (4,0.29)
        };
        \addplot[thick, draw=colorDeletions, fill=colorDeletions!32] coordinates {
            (0,0.07)
            (1,0.13)
            (2,0.25)
            (3,0.16)
            (4,0.40)
        };
    \end{axis}
\end{tikzpicture}
\iffalse\end{tikzpicture}\fi % Workaround to not confuse Overleaf's syntax highlight.
        \\[0.05cm]\textbf{(c)}
        % statistics (additions)
        %   mean: 364.3623
        %   std: 626.0631
        %   std_mean: 0.0054
        %   1%-quantile: 1
        %   2.5%-quantile: 1
        %   25%-quantile: 1
        %   median: 8
        %   75%-quantile: 440
        %   97.5%-quantile: 2265
        %   99%-quantile: 2742
        %   welford mean: 31418399.5822
        %   welford std: 54103208.5266
        % statistics (deletions)
        %   mean: 397.3647
        %   std: 450.0181
        %   std_mean: 0.0128
        %   1%-quantile: 1
        %   2.5%-quantile: 1
        %   25%-quantile: 56
        %   median: 236
        %   75%-quantile: 619
        %   97.5%-quantile: 1638
        %   99%-quantile: 2043
        %   welford mean: 34287218.1760
        %   welford std: 38883350.2997
    \end{minipage}
    \hfill
    \begin{minipage}[b]{0.4\textwidth}
        \centering
        \refstepcounter{subfigure}\label{fig:num-dels-per-triple}
        \definecolor{colorDeletions}{HTML}{D95F02}
\ifstandalone
    \begin{tikzpicture}
\else
    \begin{tikzpicture}[baseline, trim axis left]
\fi
    \begin{semilogyaxis}[
        ybar,
        scale only axis,
        width=\ifstandalone0.4\fi\textwidth,
        height=2.75cm,
        bar width=0.9cm,
        enlarge x limits=0.2,
        xlabel={Number of deletions of same triple\vphantom{[}},
        xtick={0,...,3},
        xticklabels={
            0\vphantom{[},
            1\vphantom{[},
            2--9\vphantom{[},
            $\geq$\,10\vphantom{[}
        },
        xtick pos=bottom,
        enlarge y limits={upper, value=0.2},
        ylabel={Frequency},
        ylabel near ticks,
        yticklabel pos=right,
        ytick style={draw=none},
        ymin=1,
        ymajorgrids,
        max space between ticks=15,
    ]
        \addplot[thick, draw=colorDeletions, fill=colorDeletions!33] coordinates {
            (0,12208588961)
            (1,1217826963)
            (2,8778075)
            (3,214630)
        };
        \draw[anchor=south, font=\footnotesize]
            (axis cs:0,12208588961) node {$1.2\!\cdot\!10^{10}$}
            (axis cs:1,1217826963) node {$1.2\!\cdot\!10^9$}
            (axis cs:2,8778075) node {$8.8\!\cdot\!10^6$}
            (axis cs:3,214630) node {$2.1\!\cdot\!10^5$}
        ;
    \end{semilogyaxis}
\end{tikzpicture}
\iffalse\end{tikzpicture}\fi % Workaround to not confuse Overleaf's syntax highlight.
        \\[0.05cm]\textbf{(d)}
        % mean: 0.0927
        % std: 0.3604
        % std_mean: 0.0000
        % 1%-quantile: 0
        % 2.5%-quantile: 0
        % 25%-quantile: 0
        % median: 0
        % 75%-quantile: 0
        % 97.5%-quantile: 1
        % 99%-quantile: 1
    \end{minipage}
    \addtocounter{figure}{-1}
    \caption{
        \textbf{(a)}~Histogram of time between consecutive revisions of the same entity (mean 69.98 days, standard deviation 149.39 days, median 12~days).
        \textbf{(b)}~Histogram of number of RDF triple additions per revision (mean~9.56, standard deviation~38.60, median~2) and deletions per revision (mean~0.88, standard deviation~14.73, median~0).
        \textbf{(c)}~Histogram of time until a RDF triple is first added/deleted.
        For added triples, the time since the creation of the entity is measured (mean 363.64 days, standard deviation 626.19 days, median 8~days).
        For deleted triples, the time since the triple has been added is measured (mean 396.84 days, standard deviation 450.04 days, median 236~days).
        \textbf{(d)}~Histogram of number of deletions of same RDF triple (mean~0.09, standard deviation~0.36, median~0).}
\end{figure}
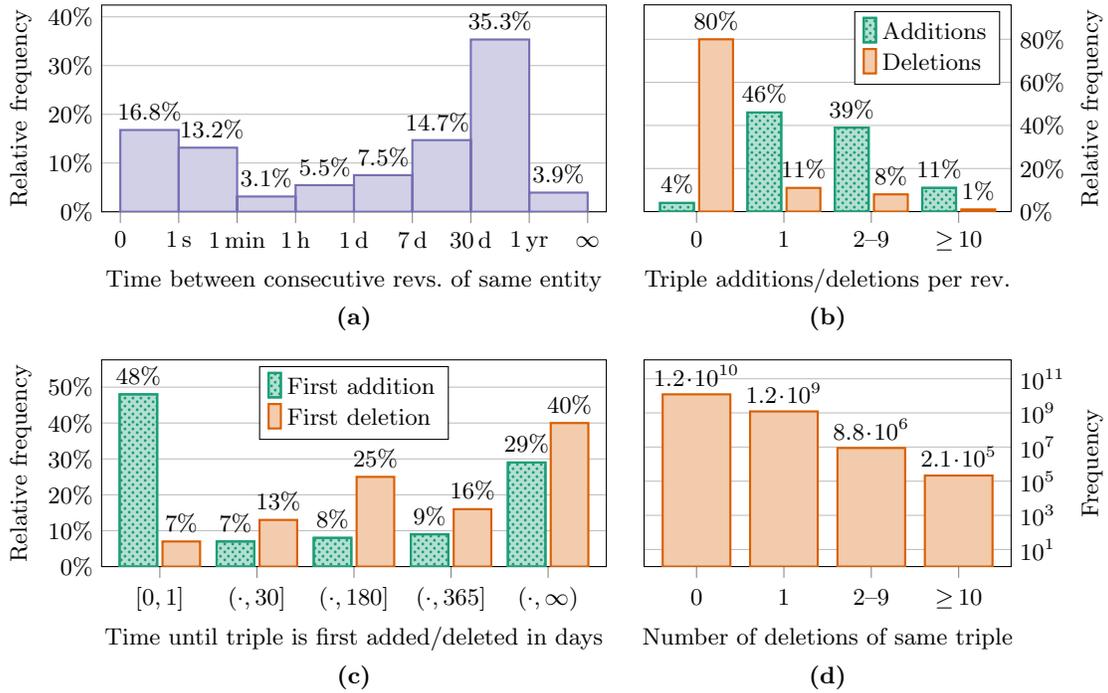

The time between consecutive revisions of the same entity is visualized in \cref{fig:time-between-revisions}.
30\% of revisions are being authored within less than a minute since the previous revision.
We suspect that this stems either from heavy activity on the most popular entities, or from editors performing multiple related changes directly after another, such as changing two related statements or reverting erroneous edits.
For 60\% of revisions, the time since the previous revision is less than a month.
For less than 4\% of revisions, that time is more than a year.
Coupled with the previous figure's data of only 2.5\% of entities having exactly one revision, it stands to reason that most entities in Wikidata are edited somewhat frequently.
However, this does not imply that most entities are checked by humans with some frequency, as these changes could also have been made by automated bots.
\idea{Figure/data point for this?}

Getting closer to the contents of revisions, we look at the number of additions and deletions of RDF triples per revision in \cref{fig:num-triple-adds-dels-per-revision}.
Note that there is no one-to-one correspondence between the number of Wikidata statements and the number of RDF triples.
For example, a single site link is expressed in more than one triple (compare \crefrange{line:q42rdf-sitelink-start}{line:q42rdf-sitelink-end} of \cref{lst:q42rdf}).
Additions are much more common than deletions with 80\% of revisions not featuring any triple deletions.
Since 89\% of revisions contain less than 10~triple additions, we conclude that most revisions constitute atomic changes and that the case in which multiple statements of a single entity change is much rarer.

In \cref{fig:time-until-triple-add-del}, we show the time until a RDF triple is first added or deleted.
Approximately half of triples are added within less than a day since the the creation of its entity.
Deletions take far longer: more than half of all deleted triple are deleted more than half a year after they had originally been added.
Besides changes to the Wikidata schema---like the deletion of properties---that potentially entail (semi-) automated changes to otherwise unchanged entities, we see two competing explanations for these late deletions: it might simply take a while until facts in the real world change and Wikidata can only update its record of them once they do, or alternatively, Wikidata might take a while to detect incorrect knowledge for the less popular entities.
A more detailed look into classifying the types and causes of changes will therefore be necessary for further investigation.

Last, \cref{fig:num-dels-per-triple} visualizes repeated deletions of the same RDF triple.
Unsurprisingly, the vast majority of triples added to Wikidata are never deleted.
Slightly less than 10\% of triples are deleted exactly one time; 4\% of which are added back into Wikidata again afterwards (not shown in figure).
Even though only less than 1\% of triples are deleted from Wikidata more than once, a few of these are deleted very many times.
For example, around 52 thousand triples are deleted and added to Wikidata more than 100 times (not shown in figure).
We suspect heavy edit wars---potentially between bots---as the main cause for this.
\idea{Which type of triples is deleted most often?}

To summarize, we have quantified how Wikidata changes over time on a macro level through analyzing statistical characteristics of \wikidated, which demonstrates its fitness as a dataset for evolving knowledge graph research.
    \section{Related Work}
\label{sec:related}

Based on their naming, Wikidata's incremental dumps\footnote{Available at \url{https://dumps.wikimedia.org/other/incr/wikidatawiki/}.} may seem to address the exact same problem as \wikidated.
These dumps are published every 24 hours and contain all revisions authored since the last dump.
However, like the full dumps discussed in \cref{sec:data-model}, each revision is stored in its full state and no obvious way exists to identify what changed from one revision to the next.
Additionally, incremental dumps older than a few months are routinely taken offline.
Because of this, they do not offer a way to trace the full edit history since Wikidata's inception like \wikidated\ does.
Their main use case is to keep live services operating on Wikidata's contents up-to-date.

Much closer to our setting is the history query service\footnote{Available at \url{https://wdhqs.wmflabs.org/}, however only displaying a \enquote{502 Bad Gateway} error during the time of writing (June to October 2021). Documentation at \url{https://www.wikidata.org/wiki/Wikidata:History_Query_Service}.}~\cite{PellissierTanonSuchanek2019QueryingEditHistory}.
It consists of a SPARQL~\cite{TheW3CSPARQLWorkingGroup2013SPARQLOverview} endpoint that allows users to query for Wikidata revision differences---similarly to \wikidated.
The paper's main contributions are on how to express revision additions and deletions in a RDF data model and how to index them for efficient query answering.
On the other hand, the paper does not discuss \emph{how} revision additions and deletions are computed, does not discuss any limitations, and does not provide a stable, downloadable dataset.
It is therefore not suitable as an environment for reproducible evaluations.

In contemporary work, Shenoy et al.~\cite{ShenoyEtAl2021StudyQualityWikidata} follow an alternative approach\footnote{Data and analysis scripts available at \url{https://w3id.org/wd_quality}.} for studying changes in Wikidata over time.
In contrast to our approach of parsing a single dump of Wikidata's full revision history, they utilize monthly dumps of Wikidata's current state, i.\,e., dumps that only contain the most recent revision of each entity for the respective month.
They then analyze which statements were deleted and added from one month to the next.
In comparison to \wikidated, which records statement changes at the revision level and thus includes revision metadata such as the exact point in time when a statement was deleted/added, their approach thus only aggregates all changes to an entity per month.
This aggregation implies the inability to record phenomena such as the frequency of revisions to entities per month or changes that are reverted within the same month.
The upside of their approach is that records of deleted entities, which are purged from the full revision history dump, are still available in those monthly dumps that were created before the entity was deleted.
Their data thus provides a useful addition to \wikidated.

The CorHist dataset~\cite{PellissierTanonEtAl2019LearningHowCorrect} is another dataset build from Wikidata's edit history.
However, it limits itself to recording constraint-related data.
Wikidata constraints are similar to database integrity constraints and used to aid Wikidata editors in finding erroneous data.
Specifically, the CorHist dataset records past constraint violations and their corrections.
\wikidated, in comparison, records all statement changes (including constraint violations, albeit in a different format than CorHist and only implicitly) including revision metadata, such as when and by whom a revision was authored, which makes it a more complete resource.

Other research that studies Wikidata's evolution includes Sarasua et al.~\cite{SarasuaEtAl2019EvolutionPowerStandard}, which studies the engagement of Wikidata's editors over time; Piscopo et al.~\cite{PiscopoEtAl2017ProvenanceInformationCollaborative}, which evaluates the quality of provenance information in Wikidata; and Piscopo and Simperl~\cite{PiscopoSimperl2018WhoModelsWorld}, which investigates the relation been different types of editors and their impact on the Wikidata ontology over time.
    \section{Conclusion}
\label{sec:concl}

We have presented \wikidated, a dataset containing Wikidata's revision history as incremental revisions, i.\,e., sets of deletions and additions of RDF triples.
To the best of our knowledge, it constitutes the first large evolving knowledge graph dataset of its kind.
We foresee applications both from the Wikidata community for studying how Wikidata changed over time, as well as from the wider knowledge graph community for evaluating techniques over evolving knowledge graphs, such as incremental knowledge graph embeddings or updatable indexing structures for efficient query answering.

Besides releasing the dataset and the accompanying codebase\cref{note:download}, in this paper we have documented our methodology for creating \wikidated, and discussed its implementation and limitations, the biggest one being the omission of deleted entities, which are not contained in the openly available revision history dumps of Wikidata.
Additionally, we have presented statistical characteristics of our dataset, and compared it to related work.

\subsection{Future Work}
\label{sec:future}

In the future, we plan to release new versions and additional variants of our dataset.
In particular, we aim to establish a release cadence for publishing a new Wikidated version on the most recent Wikidata dump in regular time intervals, based on community uptake.
Additionally, we are working on extracting subsets of \wikidated\ that only contain RDF serializations of simple statements, i.\,e., statements with qualifiers and references removed, or \enquote{thruthy} statements, i.\,e., statements with the highest rank---similar to the existing equally-named variant of the official Wikidata dumps.
Last, we are thinking about additional ways of reducing the large file size of \wikidated, such as switching to better compression formats, subsampling the dataset, or aggregating deletions and additions of all revisions in a fixed time frame (e.\,g., an hour, day, or week).

Further future work includes deeper analysis of the editing dynamics of Wikidata recorded in \wikidated, such as detecting updates from sets of triple deletions and additions, or classifying the type and source of changes; and consideration of Wikidata's more recent lexicographic data, the first step towards this would be to implement a RDF serialization of it in Wikidata Toolkit.
We invite the Wikimedia Foundation specifically to consider also releasing official Wikidata dumps in an incremental format---such as the one used for \wikidated---in order to save bandwidth and storage space for users.
Additionally, we would welcome any way to access and integrate revision data of deleted entities and revisions into Wikidated.
While the raw data itself contains sensible personal information and copyrighted material that is unavailable to the general public for good reason, the metadata of deleted entities and revisions, e.\,g., number of deleted statements, is by itself interesting to us.
For instance, it could be published by replacing all literal values in deleted revisions with generated ones, thus only preserving the data's graph structure.

\subsubsection*{Acknowledgments}

Lukas Schmelzeisen was supported by the German Research Foundation~(DFG) via grant agreement number STA 572/18-1~(Open Argument Mining).
Corina Dima was supported by the German Federal Ministry for Economic Affairs and Energy~(BMWi) via grant agreement number 01MK20008F~(Service-Meister).
Thanks to Raphael Menges for suggesting the name \enquote{Wikidated}.

    \bibliographystyle{splncs04}
    \bibliography{wikidated}
\end{document}